\pdfoutput=1
\documentclass[11pt]{article}

\usepackage{acl}

\usepackage{times}
\usepackage{latexsym}

\usepackage{hyperref}
\usepackage{url}
\usepackage{graphicx}
\PassOptionsToPackage {dvipsnames}{xcolor}
\usepackage{tikz}
\usepackage{caption}
\usepackage{subcaption}
\usepackage{wrapfig}
\usepackage{MnSymbol}
\usepackage{algorithm}
\usepackage{algorithmic}
\usepackage{booktabs, multirow}
\usepackage{makecell}
\usepackage{amsmath}
\usepackage{mathtools}
\usepackage{amsfonts}
\usepackage{pifont}
\usepackage{ntheorem}
\usepackage{ragged2e,array}
\usepackage{sidecap}
\usepackage{multicol}
\usepackage{CJKutf8}
\usepackage{commath}
\usepackage{mathtools}
\sidecaptionvpos{figure}{c}
\usepackage{times}

\usepackage{enumitem}

\newlist{questions}{enumerate}{2}
\setlist[questions,1]{label=\textbf{RQ\arabic*:},ref=RQ\arabic*}
\setlist[questions,2]{label=(\alph*),ref=\thequestionsi(\alph*)}

\definecolor{brewer-blue}{RGB}{30, 100, 220}
\definecolor{brewer-red}{RGB}{220,0,0}

\usetikzlibrary{shapes,decorations,arrows,calc,arrows.meta,fit,positioning}
\tikzset{
    -Latex,auto,node distance =1 cm and 1 cm,semithick,
    state/.style ={ellipse, draw, minimum width = 0.7 cm},
    point/.style = {circle, draw, inner sep=0.04cm,fill,node contents={}},
    bidirected/.style={Latex-Latex,dashed},
    el/.style = {inner sep=2pt, align=left, sloped}
}

\theoremseparator{:}
\newtheorem{hyp}{Hypothesis}

\usepackage[T1]{fontenc}
\usepackage[utf8]{inputenc}

\usepackage{microtype}

\title{Interpreting the Robustness of Neural NLP Models to Textual Perturbations}

\author{
Yunxiang Zhang$^1$,
Liangming Pan$^2$,
Samson Tan$^2$,
Min-Yen Kan$^2$
\\
$^1$Wangxuan Institute of Computer Technology, Peking University\\
$^2$School of Computing, National University of Singapore\\
\\
{\tt yx.zhang@pku.edu.cn,} \
{\tt liangmingpan@u.nus.edu,} \\
{\tt \{samson.tmr,kanmy\}@comp.nus.edu.sg} \
}

\begin{document}
\maketitle
\begin{abstract}
Modern Natural Language Processing (NLP) models are known to be sensitive to input perturbations and their performance can decrease when applied to real-world, noisy data.  However, it is still unclear why models are less robust to some perturbations than others. In this work, we test the hypothesis that the extent to which a model is affected by an unseen textual perturbation (robustness) can be explained by the learnability of the perturbation (defined as how well the model learns to identify the perturbation with a small amount of evidence). We further give a causal justification for the learnability metric. We conduct extensive experiments with four prominent NLP models --- TextRNN, BERT, RoBERTa and XLNet --- over eight types of textual perturbations on three datasets. We show that a model which is better at identifying a perturbation (higher learnability) becomes worse at ignoring such a perturbation at test time (lower robustness), providing empirical support for our hypothesis.
\end{abstract}

\section{Introduction}\label{sec:intro}

Despite the success of deep neural models on many Natural Language Processing (NLP) tasks~\citep{liu2016recurrent, devlin2019bert, liu2019roberta}, recent work has discovered that these models are not robust to noisy input from the real world and thus their performance will  decrease ~\citep{prabhakaran2019perturbation,niu2020evaluating,ribeiro2020beyond,moradi2021evaluating}. A reliable NLP system should not be easily fooled by slight noise in the text. Although a wide range of evaluation approaches for robust NLP models have been proposed~\citep{ribeiro2020beyond,morris-etal-2020-textattack,goel2021robustness,wang-etal-2021-textflint}, few attempts have been made to \textit{understand} these benchmark results. Given the difference of robustness  between models and perturbations, it is a natural question why models are more sensitive to some perturbations than others. It is crucial to avoid over-sensitivity to input perturbations, and understanding why it happens is useful for revealing the weaknesses of current models and designing more robust training methods. To the best of our knowledge, a quantitative measure to \textit{interpret} the robustness of NLP models to textual perturbations has yet to be proposed.
% \begin{figure}
%     \centering
%     \includegraphics[width=0.45\textwidth]{robustness_data_aug_causal_estimate.pdf}
%      \caption{Robustness vs. post data augmentation  $\Delta$ vs. average learnability  on IMDB dataset. Each point in the plots represents a model-perturbation pair. 
%      We define ``robustness'' as the performance drop on perturbed test set, ``post aug $\Delta$'' as the performance boost on perturbed test set after data augmentation along such a perturbation, and
%      ``average learnability'' as how well the model learns to identify the perturbation with a small amount of evidence.
%      }\label{fig:findings}
% \end{figure}
To improve the robustness under perturbation, it is common practice to leverage data augmentation~\citep{li-specia-2019-improving,min2020syntactic,tan-joty-2021-code}. Similarly, how much data augmentation through the perturbation improves model robustness varies between models and perturbations. In this work, we aim to investigate two Research Questions (RQ): 
\begin{itemize}
    \item \textbf{RQ1:} \textit{Why are NLP models less robust to some perturbations than others?}\label{itm:rq1}
    \item \textbf{RQ2:} \textit{Why does data augmentation work better at improving the model robustness to some perturbations than others?}\label{itm:rq2}
\end{itemize}
% \begin{questions}
%         \vspace{-0.2cm}\item \textit{Why NLP models are less robust to some perturbations than others?}\label{itm:rq1}
%         \vspace{-0.2cm}\item \textit{Why data augmentation works better at improving the model robustness to some perturbations than others?}\label{itm:rq2}
% \end{questions}

\begin{table*}
\renewcommand\arraystretch{1.2}
\centering
\begin{tabular}{clccccc}\toprule
Exp No. & Measurement &  Label & Perturbation & Training Examples & Test Examples\\\cmidrule{1-6}
0 &Standard &original & $l \in \varnothing$&$(x_{i},0), (x_{j}, 1)$ & $(x_{i},0), (x_{j}, 1)$\\\cmidrule{1-6}
1 &Robustness &original &$l \in \{0,1\}$ & $(x_{i},0), (x_{j}, 1)$ &$ (x_{i}^{*},0), (x_{j}^{*}, 1)$ \\\cmidrule{1-6}
2 &Data Augmentation &original & $l \in \{0,1\}$&\makecell[c]{$(x_{i},0), (x_{j}, 1)$ \\ $(x_{i}^{*},0), (x_{j}^{*}, 1)$} & $(x_{i}^{*},0), (x_{j}^{*}, 1)$\\\cmidrule{1-6}
3 &\multirow{2}{*}{Learnability} &random &$l^{\prime} \in \{1^{\prime}\}$ & $(x_{j},0^{\prime}), (x_{i}^{*}, 1^{\prime})$ &$ (x_{i}^{*}, 1^{\prime})$ \\
4& &random & $l^{\prime} \in \{1^{\prime}\}$&$(x_{j},0^{\prime}), (x_{i}^{*}, 1^{\prime})$ & $(x_{i}, 1^{\prime})$\\
\bottomrule
\end{tabular}
\caption{Example experiment settings for measuring learnability, robustness and improvement by data augmentation. We perturb an example if its label falls in the set of label(s) in ``Perturbation'' column. $\varnothing$ means no perturbation at all. Training/test examples  are the expected input data,  assuming we have only  one negative  $(x_{i},0)$ and  positive  $(x_{j}, 1)$ example in our original training/test set. $l^{\prime}$  is a random label and $x^{*}$  is a perturbed example.  }\label{tab:settings}
\end{table*}

We test a hypothesis for RQ1 that the extent to which a model is affected by an unseen textual perturbation (robustness) can be explained by the learnability of the perturbation (defined as how well the model learns to identify the perturbation with a small amount of evidence). We also validate another hypothesis for RQ2 that the learnability metric is predictive of the improvement on robust performance brought by data augmentation along a perturbation. Our proposed learnability is inspired by the concepts of Randomized Controlled Trial (RCT) and Average Treatment Effect (ATE) from Causal Inference~\citep{rubin1974estimating, holland1986statistics}. Estimation of perturbation learnability for a model consists of three steps:   \ding{172} randomly labelling a dataset, \ding{173} perturbing examples of a particular pseudo class with probabilities, and \ding{174} using ATE to measure the ease with which the model learns the perturbation. The core intuition for our method is to frame an RCT as a perturbation identification task and formalize the notion of learnability as a causal estimand based on ATE. We conduct extensive experiments on four neural NLP models with eight different perturbations across three datasets and find strong evidence for our two hypotheses. Combining these two findings, we further show that data augmentation is \textit{only} more effective at improving robustness against perturbations that a model is more sensitive to, contributing to the interpretation of robustness and data augmentation. Learnability provides a clean setup for analysis of the  model behaviour under perturbation, which contributes better model interpretation as well.

\paragraph{Contribution.} This work provides an empirical explanation for why NLP models are less robust to some perturbations than others. The key to this question is perturbation learnability, which is grounded in the causality framework. We show a statistically significant inverse correlation between learnability and robustness. 

\section{Setup and Terminology}\label{sec:setup}

As a pilot study, we consider the task of binary text classification. The training set is denoted as $D_{train} = \{(x_{1}, l_{1}), ...,  (x_{n}, l_{n})\}$, where  $x_{i}$ is the $i$-th example and $l_{i} \in \{0,1\}$ is the corresponding label. We fit a model $f : (x;\theta) \mapsto \{0,1\}$ with parameters $\theta$ on the training data. A textual perturbation is a transformation $g : (x;\beta) \rightarrow x^{*} $ that injects a specific type of noise into an example $x$ with parameters $\beta$ and the resulting perturbed example is $x^{*}$. We design several experiment settings (Table~\ref{tab:settings}) to answer our research questions.  Experiment 0 in Table~\ref{tab:settings} is the  standard learning  setup, where we train and evaluate a model on the original dataset. Below we detail other experiment settings.

\subsection{Definitions}\label{sec:defi}

\paragraph{Robustness.}

We apply the perturbations to test examples and measure the robustness of model to said perturbations as the  decrease in accuracy. In Table~\ref{tab:settings}, Experiment 1 is related to robustness measurement, where we train a model on unperturbed dataset and test it on perturbed examples. We denote the test accuracy  of a model $f(\cdot)$  on  examples perturbed by $g(\cdot)$ in Experiment 1 as $\mathcal{A}_{1}(f,g,D^{*}_{test})$. Similarly, the test accuracy in Experiment 0 is $\mathcal{A}_{0}(f,D_{test})$.  Consequently, the robustness is calculated as the difference of test accuracies:
\begin{equation}
\begin{split}
    \text{robustness}(f,g,D)=\mathcal{A}_{1}(f,g,D^{*}_{test})\\
    -\mathcal{A}_{0}(f,D_{test}).
\end{split}
\label{eq:robustness}
\end{equation}
Models usually suffer a performance drop when encountering perturbations, therefore the robustness is usually negative, where lower values indicate decreased robustness.

\paragraph{Improvement by Data Augmentation (Post Augmentation $\Delta$).}
To improve robust accuracy~\citep{tu2020empirical} (i.e., accuracy on the perturbed test set), it is a common practice to leverage data augmentation~\citep{li-specia-2019-improving,min2020syntactic,tan-joty-2021-code}. We simulate the data augmentation process by appending perturbed data to the training set (Experiment 2 of Table~\ref{tab:settings}). We calculate the improvement on performance after data augmentation as the difference of test accuracies:
\begin{equation}
\begin{split}
        \Delta_{\text{post\_aug}}(f,g,D) = \mathcal{A}_{2}(f,g,D^{*}_{test})\\
        -\mathcal{A}_{1}(f,g,D^{*}_{test}).
\end{split}
\label{eq:data-aug}
\end{equation}
where $\mathcal{A}_{2}(f,g,D^{*}_{test})$ denotes the test accuracy of Experiment 2. 
$\Delta_{\text{post\_aug}}$ is the higher the better.

\paragraph{Learnability.}
We want to compare perturbations in terms of how well the model \textit{learns} to identify them with a small amount of evidence. We cast learnability estimation as a perturbation classification task, where a model is trained to identify the perturbation in an example. We define that the learnability estimation consists of three steps, namely \ding{172} \textbf{assigning random labels}, \ding{173} \textbf{perturbing with probabilities}, and \ding{174} \textbf{estimating model performance}. Below we introduce the procedure and intuition for each step. This estimation framework is further grounded in concepts from the causality literature in Section~\ref{sec:causal}, which justifies our motivations. We summarize our estimation approach formally in Algorithm~\ref{alg:algorithm} (Appendix~\ref{sec:algo}). 

\noindent \textbf{\ding{172} Assigning Random Labels.} We randomly assign pseudo labels to each training example regardless of its original label. Each data point has equal probability of being assigned to positive ($l^{\prime}=1$) or negative ($l^{\prime}=0$) pseudo label. This results in a randomly labeled dataset $D_{train}^{\prime} = \{(x_{1}; l_{1}^{\prime}), ...,  (x_{n}, l_{n}^{\prime})\}$, where $L^{\prime} \sim Bernoulli(1, 0.5)$. In this way, we ensure that there is no difference between the two pseudo groups since the data are randomly split.\label{step:1}
    
\noindent \textbf{\ding{173} Perturbing with Probabilities.} We apply the perturbation $g(\cdot)$ to each training example in one of the pseudo groups (e.g., $l^{\prime}=1$ in Algorithm~\ref{alg:algorithm})\footnote{Because the training data is randomly split into two pseudo groups, applying perturbations to any one of the groups should yield same result. We assume that we always perturb into the first group ($l^{\prime}=1$) hereafter.}. In this way, we create a correlation between the existence of perturbation and label (i.e., the perturbation occurrence is predictive of the label). We control the perturbation probability $p \in [0,1] $,  i.e., an example has a specific probability $p$ of being perturbed. This results in a perturbed training set $D_{train}^{\prime *} = \{(x_{1}^{*}, l_{1}^{\prime}), ...,  (x_{n}^{*}, l_{n}^{\prime})\}$, where the perturbed example $x_{i}^{*}$ is:
\begin{equation}
\begin{split}
    Z \sim U(0,1), \forall i \in \{1,2,...,n\}\\
    x_{i}^{*} =
    \begin{cases}
    g(x_{i}) & {l_{i}^{\prime}=1 \wedge z<p,}  \\
    x_{i} & \text{otherwise.}
    \end{cases}
\end{split}
\end{equation}
Here $Z$ is a random variable drawn from a uniform distribution $U(0,1)$. Due to randomization in the formal step, now the only difference between the two pseudo groups is the occurrence of perturbation.\label{step:2}

\noindent \textbf{\ding{174} Estimating Model Performance.} We train a model on the randomly labeled dataset with perturbed examples. Since the only difference between the two pseudo groups is the existence of the perturbation, the model is trained to identify the perturbation. The original test examples $D_{test}$  are also assigned random labels and become $D_{test}^{\prime}$. We perturb all of the test examples in one  pseudo group (e.g., $l^{\prime}=1$,  as in step~\ref{step:1}) to produce a perturbed test set $D_{test}^{\prime *}$. Finally, the perturbation learnability is calculated as the difference of accuracies on $D_{test}^{\prime *}$ and $D_{test}^{\prime}$, which indicates how much the model learns from the perturbation's co-occurrence with pseudo label:
 \begin{equation}
\begin{split}
        \text{learnability}(f,g,p,D)=\mathcal{A}_{3}(f,g,p,D^{\prime *}_{test})\\
        -\mathcal{A}_{4}(f,g,p,D^{\prime}_{test}).
\end{split}
\label{eq:learnability}
\end{equation}
 \label{step:3}
$\mathcal{A}_{4}(f,g,p,D^{\prime *}_{test})$ and $\mathcal{A}_{3}(f,g,p,D^{\prime}_{test})$ are accuracies measured by Experiment 4 and 3 of Table~\ref{tab:settings}, respectively. 

We observe that the learnability depends on perturbation probability $p$. For each model--perturbation pair, we obtain multiple learnability estimates by varying the perturbation probability (Figure~\ref{fig:sensitivity}). However, we expect that learnability of the perturbation (as a concept) should be independent of perturbation probability. To this end, we use the $\log AUC$ (area under the curve in log scale) of the $p-\text{learnability}$ curve (Figure~\ref{fig:sensitivity}), termed as ``average learnability'', which summarizes the overall learnability across different perturbation probabilities $p_{1}, ..., p_{t}$:
\begin{equation}
\begin{split}
    \text{avg\_learnability}(f,g,D) \coloneqq \log AUC(\{(p_{i}, \\\text{learnability}(f,g,p_{i},D)) \mid i \in \{1,2,...,t\}\}).
    \label{eq:logauc}
\end{split}
\end{equation}
We use $\log AUC$ rather than $AUC$ because we empirically find that the learnability varies substantially between perturbations when $p$ is small, and a log scale can better capture this nuance. We also introduce learnability at a specific perturbation probability (Learnability @ $p$) as an alternate summary metric and provide a comparison of this metric against $\log AUC$ in Appendix~\ref{sec:learnability-p}.

\subsection{Hypothesis}\label{sec:hypo}
With the above-defined terminologies, we propose hypotheses for RQ1 and RQ2 in Section \ref{sec:intro}, respectively.
\begin{hyp}[H\ref{hyp:robustness}] \label{hyp:robustness}
A model for which a perturbation is more learnable is less robust against the same perturbation at the test time.
\end{hyp}
This is \textit{not} obvious because the model encounters this perturbation during training in learnability estimation while they do not in robustness measurement.
\begin{hyp}[H\ref{hyp:data-aug}] \label{hyp:data-aug}
A model for which a perturbation is more learnable  experiences bigger robustness gains with data augmentation along such a perturbation.
\end{hyp}
We validate both Hypotheses~\ref{hyp:robustness} and~\ref{hyp:data-aug} with experiments on several perturbations and models described in Section~\ref{sec:pert} and~\ref{sec:exp-setting}.

\section{A Causal View on Perturbation Learnability}\label{sec:causal}
In Section~\ref{sec:defi}, we introduce the term ``learnability'' in an intuitive way. Now we map it to a formal, quantitative measure in standard statistical frameworks. Learnability is actually motivated by concepts from the causality literature. We provide a brief introduction to basic concepts of causal inference in Appendix~\ref{sec:background}. In fact, learnability is the causal effect of perturbation on models, which is often difficult to measure due to the confounding latent features. In the language of causality, this is ``correlation is not causation''. Causality provides insight on how to fully decouple the effect of perturbation and other latent features. We introduce the causal motivations for step \ref{step:1} and \ref{step:3} of learnability estimation in the following Section \ref{sec:random} and \ref{sec:identify}, respectively.

\subsection{A Causal Explanation for Random Label Assignment}\label{sec:random}

\begin{figure}
    \centering
    \begin{subfigure}[b]{0.24\textwidth}
         \centering
         \includegraphics[width=\textwidth]{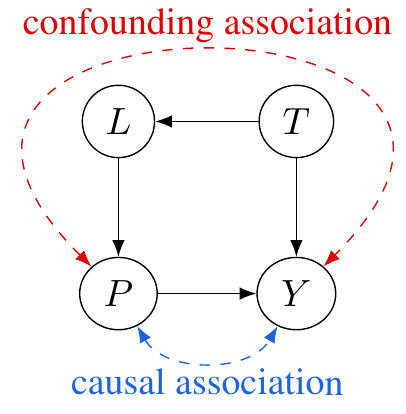}
        \caption{Before randomization.}
         \label{fig:before-random}
     \end{subfigure}
     \hfill
         \begin{subfigure}[b]{0.20\textwidth}
         \centering
          \includegraphics[width=\textwidth]{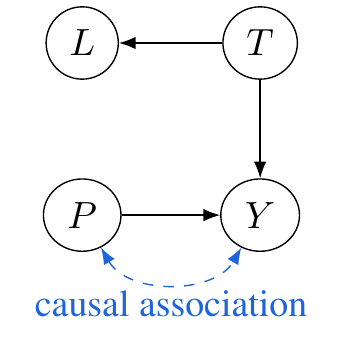}
        \caption{After randomization.}
         \label{fig:after-random}
     \end{subfigure}
         \caption{Causal graph explanation for decoupling perturbation and latent feature with randomization. $P$ is the perturbation and $T$ is the latent feature. $L$ is the original label and  $Y$ is the correctness of the predicted label.}
    \label{fig:causal-graph}
\end{figure}

Natural noise (simulated by perturbations in this work) usually co-occurs with latent features in an example. If we did not assign random labels and simply perturbed one of the \textit{original} groups, there would be confounding latent features that would prevent us from estimating the causal effect of the perturbation.  Figure~\ref{fig:before-random}  illustrates this scenario. Both perturbation $P$ and latent feature $T$  may affect the outcome $Y$,\footnote{$Y$ is later defined in Section~\ref{sec:identify}} while the latent feature is predictive of label $L$. Since we make the perturbation $P$ on examples with the same label, $P$ is decided by $L$. It therefore follows that $T$ is a confounder of the effect of $P$ on $Y$, resulting in non-causal association flowing along the path $P \leftarrow L \leftarrow T \rightarrow Y$. However, if we do randomize the labels, $P$ no longer has any causal parents (i.e., incoming edges) (Figure~\ref{fig:after-random}). This is because perturbation is purely random. Without the path represented by $P \leftarrow L$, all of the association that flows from $P$ to $Y$ is causal. As a result, we can directly calculate the causal effect from the observed outcomes.

% Our randomization experiments allow us to discern causation from association and estimate the causal effect of perturbation from test performance.

\begin{figure*}
    \centering
    \includegraphics[width=0.9\textwidth]{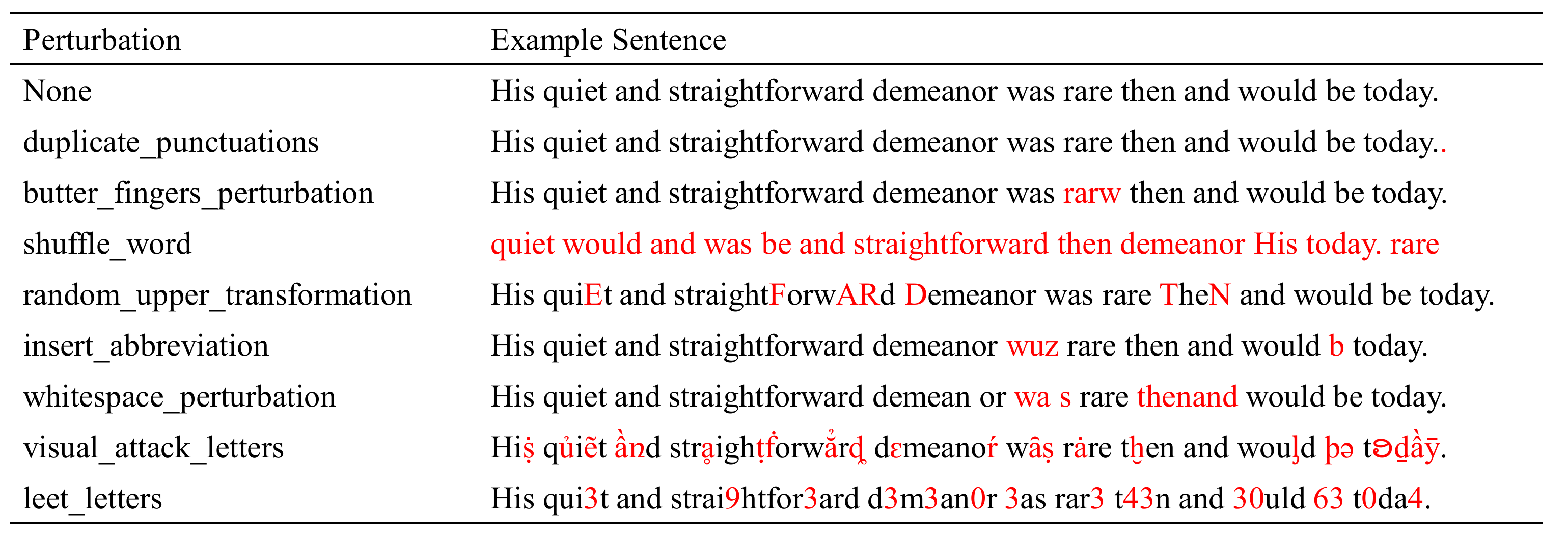}
     \caption{An example sentence with different types of perturbations.}\label{fig:example}
\end{figure*}

\subsection{Learnability is a Causal Estimand}\label{sec:identify}

We identify learnability as a causal estimand. In causality, the term ``identification'' refers to the process of moving from a causal estimand (Average Treatment Effect, ATE) to an equivalent statistical estimand. We show that the difference of accuracies on $D_{test}^{\prime *}$ and $D_{test}^{\prime}$ is actually a causal estimand. We define the outcome $Y$ of a test example $x_{i}$ as the correctness of the predicted label:
\begin{equation}
    Y_{i}(0)\coloneqq \mathbf{1}_{\{f(x_{i})=l_{i}^{\prime}\}}.
\label{eq:Y0-accu}
\end{equation}
where $\mathbf{1}_{\{\cdot\}}$ is the indicator function. Similarly, the outcome $Y$ of a perturbed test example $x_{i}^{*}$ is:
\begin{equation}
    Y_{i}(1)\coloneqq \mathbf{1}_{\{f(x_{i}^{*})=l_{i}^{\prime}\}}.
\label{eq:Y1-accu}
\end{equation}
 According to the definition of Individual Treatment Effect (ITE, see Equation~\ref{eq:ITE} of Appendix \ref{sec:background}), we have $ITE_{i} =  \mathbf{1}_{\{f(x_{i}^{*})=l_{i}^{\prime}\}} - \mathbf{1}_{\{f(x_{i})=l_{i}^{\prime}\}}$. We then take  the average over all the perturbed test examples (half of the test set)\footnote{The other half of the test set ($l^{\prime}=0$) is left unperturbed, following the same procedure in Section~\ref{sec:defi}. Model predictions will not change for unperturbed ones, resulting in ITEs with zero values. Therefore, we do not take them into account for ATE calculation.}. This is our Average Treatment Effect (ATE):
\begin{eqnarray}\label{eq:ATE-accu}
    ATE&=&\mathbb{E}[Y(1)]-\mathbb{E}[Y(0)] \nonumber \\
    &=&\mathbb{E}[\mathbf{1}_{\{f(x^{*})=l^{\prime}\}}] - \mathbb{E}[\mathbf{1}_{\{f(x)=l^{\prime}\}}] \nonumber \\ &=&P(f(x^{*})=l^{\prime})-P(f(x)=l^{\prime}) \nonumber \\
    &=&\mathcal{A}(f,g,p,D_{test}^{\prime *})- \mathcal{A}(f,g,p,D_{test}^{\prime}). \nonumber \\
\end{eqnarray}
where $\mathcal{A}(f,g,p,D)$ is the accuracy of model $f(\cdot)$ trained with perturbation $g(\cdot)$ at perturbation probability $p$ on test set $D$. Therefore, we show that ATE is exactly the difference of accuracy on the perturbed and unperturbed test sets with random labels. And the difference is learnability according to Equation \ref{eq:learnability}.

We discuss another means of identification of ATE in Appendix~\ref{sec:alter}, based on the prediction probability. We compare between the probability-based and accuracy-based metrics there. We find that our accuracy-based metric yields better resolution, so we report this metric in the main text of this paper.

\begin{figure*}
\centering
\includegraphics[width=\textwidth]{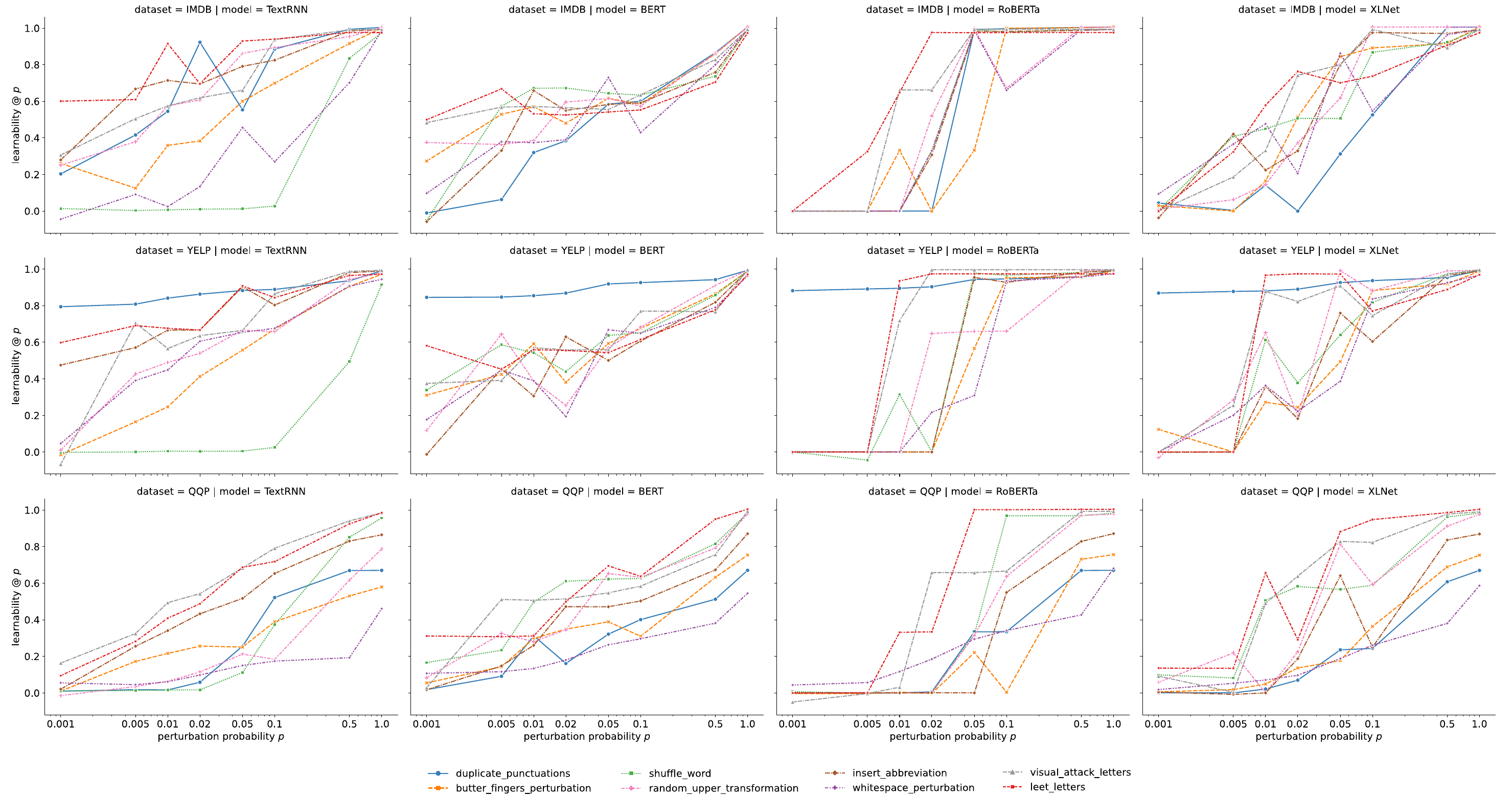}
\caption{Learnability of eight perturbations for four NLP models on three datasets, as a function of perturbation probability.}
\label{fig:sensitivity}
\end{figure*}

\begin{table*}
  \centering

\begin{tabular}{l|cccc|c}
\toprule
Perturbation & XLNet & RoBERTa & BERT  & TextRNN & \makecell[c]{Average \\ over models} \\
\midrule
whitespace\_perturbation & 1.638  & 1.436  & 1.492  & 0.878  & 1.361  \\
shuffle\_word & 1.740  & 1.597  & 1.766  & 0.594  & 1.424  \\
duplicate\_punctuations & 1.086  & 1.499  & 1.347  & 2.050  & 1.495  \\
butter\_fingers\_perturbation & 1.590  & 1.369  & 1.788  & 1.563  & 1.578  \\
random\_upper\_transformation & 1.583  & 1.520  & 1.721  & 2.039  & 1.716  \\
insert\_abbreviation & 1.783  & 1.585  & 1.564  & \underline{2.219}  & 1.788  \\
visual\_attack\_letters & \textbf{1.824} & \underline{1.921}  & \textbf{1.898 } & 2.094  & \underline{1.934}  \\
leet\_letters & \underline{1.816}  & \textbf{2.163} & \underline{1.817}  & \textbf{2.463} & \textbf{2.065} \\
% \midrule
% Average over features & 1.632  & 1.636  & 1.674  & 1.738  & 1.670  \\
\bottomrule
\end{tabular}

   \caption{Average learnability ($\log AUC$ of corresponding curve in Figure~\ref{fig:sensitivity}) of each model--perturbation pair on IMDB dataset. Rows are sorted by average values over all models. The perturbation for which a model is most learnable is highlighted in \textbf{bold} while the following one is \underline{underlined}.} \label{tab:benchmark}
\end{table*}

\begin{figure*}
     \centering
     \begin{subfigure}[b]{0.27\textwidth}
         \centering
         \includegraphics[width=\textwidth]{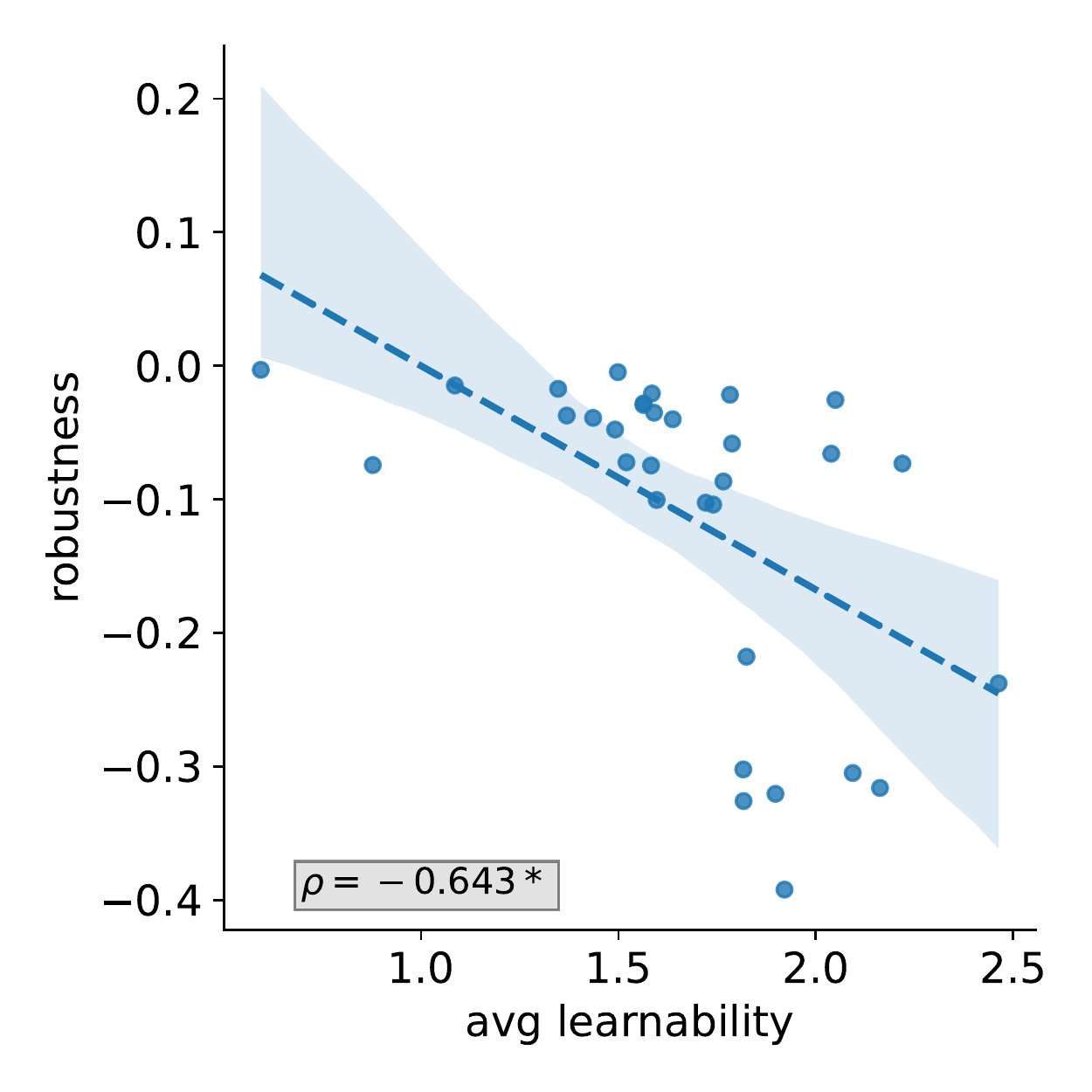}
         \caption{Learnability vs. Robustness}
         \label{fig:imdb-robustness}
     \end{subfigure}
     \hfill
     \begin{subfigure}[b]{0.27\textwidth}
         \centering
         \includegraphics[width=\textwidth]{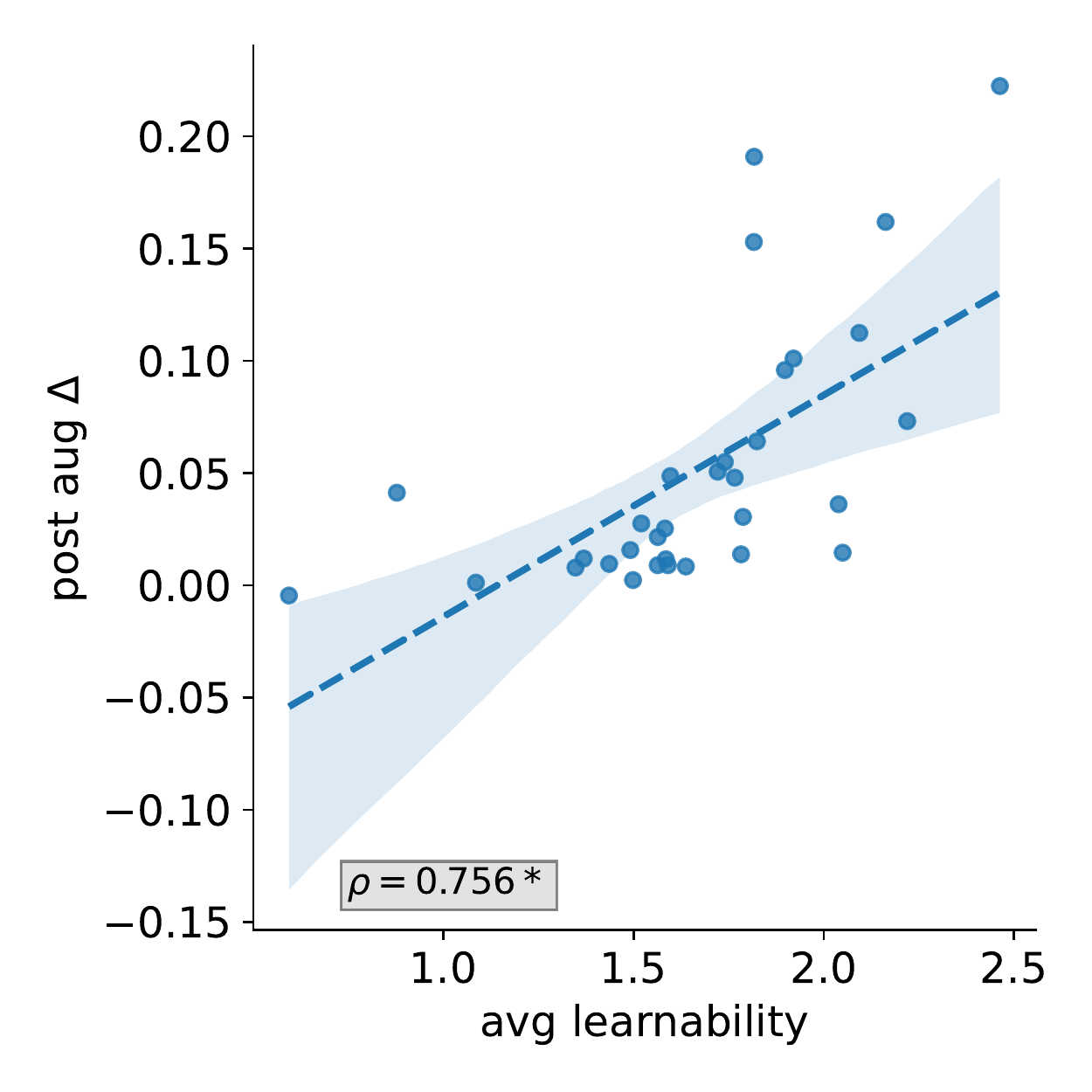}
         \caption{Learnability vs. Post Aug $\Delta$}
         \label{fig:imdb-data-aug}
     \end{subfigure}
     \hfill
     \begin{subfigure}[b]{0.38\textwidth}
         \centering
         \includegraphics[width=\textwidth]{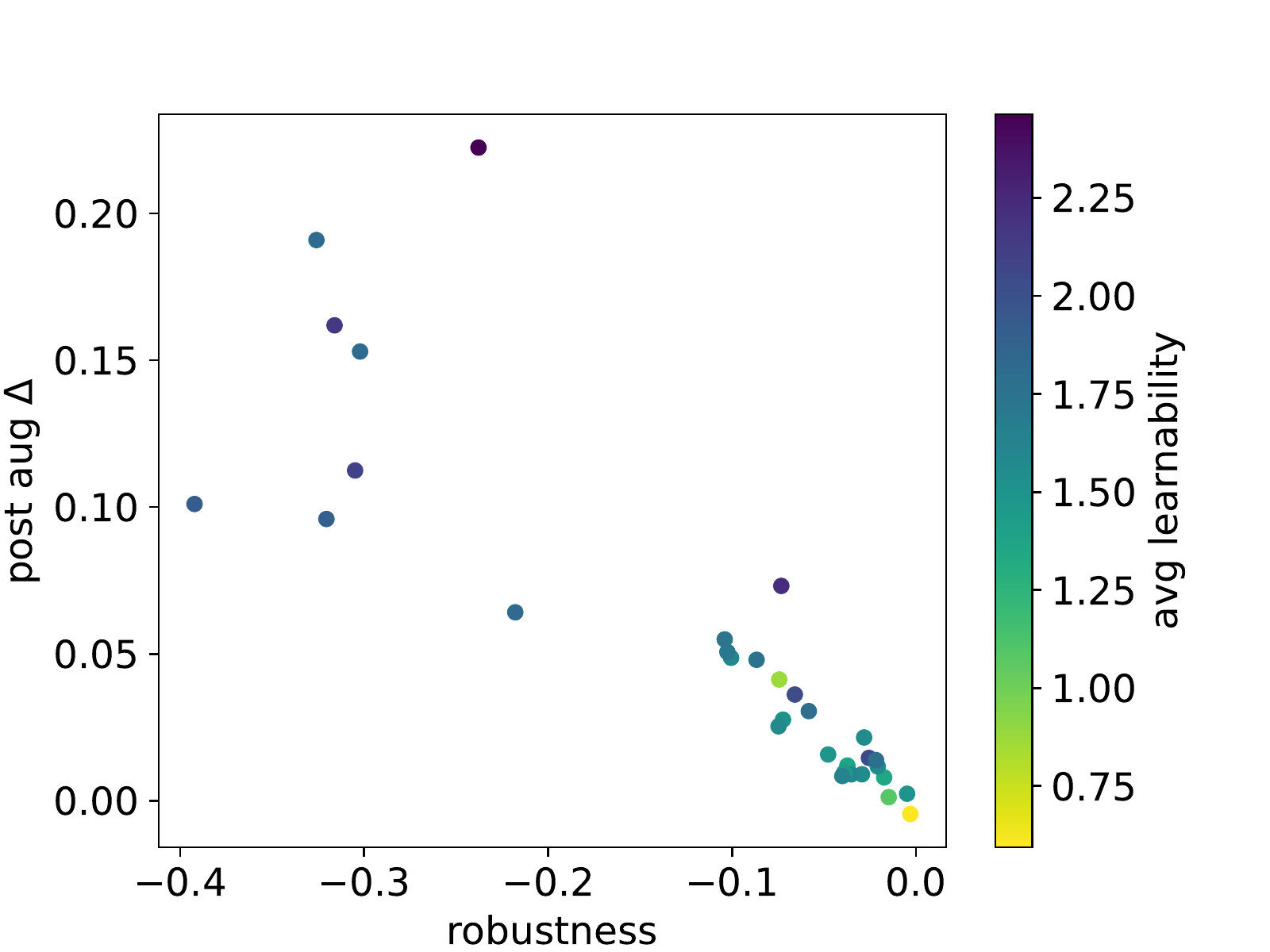}
         \caption{Learn. vs. Robu. vs. Post Aug $\Delta$}
         \label{fig:imdb-robustness-data-aug-sensitivity}
     \end{subfigure}
        \caption{Linear regression plots of learnability vs. robustness vs. post data augmentation $\Delta$  on \textbf{IMDB} dataset. Each point in the plots represents a model-perturbation pair. $\rho$ is Spearman correlation. $^{*}$ indicates high significance (p-value $<$ 0.001).
        }
        \label{fig:imdb-corr}
\end{figure*}

\section{Experiments}\label{sec:exp}

\subsection{Perturbation methods}\label{sec:pert}

\paragraph{Criteria for Perturbations.} 
We select various character-level and word-level perturbation methods in existing literature that simulate different types of noise an NLP model may encounter in real-world situations. These perturbations are non-adversarial, label-consistent, and can be automatically generated at scale. We note that our perturbations do not require access to the model internal structure. We also assume that the feature of perturbation does not exist in the original data. Not all perturbations in the existing literature are suitable for our task. For example, a perturbation that swaps gender words (i.e., female $\rightarrow$ male, male $\rightarrow$ female) is not suitable for our experiments since we cannot distinguish the perturbed text from an unperturbed one. In other words, the perturbation function $g(\cdot)$ should be \emph{asymmetric}, such that $g(g(x)) \neq x$.

Figure~\ref{fig:example} shows an example sentence with different perturbations. Perturbation of ``duplicate\_punctuation'' doubles the punctuation by appending a duplicate after each punctuation, e.g., ``,'' $\rightarrow$ ``,,'';  ``butter\_fingers\_perturbation'' misspells some words with noise erupting from keyboard typos; ``shuffle\_word'' randomly changes the order of word in the text~\citep{moradi2021evaluating}; ``random\_upper\_transformation'' randomly adds upper cased letters~\citep{wei2019eda}; ``insert\_abbreviation'' implements a rule system that encodes word sequences associated with the replaced abbreviations; ``whitespace\_perturbation'' randomly removes or adds whitespaces to text; ``visual\_attack\_letters'' replaces letters with visually similar, but different, letters~\citep{eger-etal-2019-text}; ``leet\_letters'' replaces letters with leet, a common encoding used in gaming~\citep{eger-etal-2019-text}.

\subsection{Experimental Settings}\label{sec:exp-setting}
To test the learnability, robustness and improvement by data augmentation with different NLP models and perturbations, we experiment with four modern and representative neural NLP models: TextRNN~\citep{liu2016recurrent}, BERT~\citep{devlin2019bert}, RoBERTa~\citep{liu2019roberta} and XLNet~\citep{yang2019xlnet}. For TextRNN, we use the implementation by an open-source text classification toolkit NeuralClassifier~\citep{liu-etal-2019-neuralclassifier}. For the other three pretrained models, we use the \verb|bert-base-cased|, \verb|roberta-base|, \verb|xlnet-base-cased| versions from Hugging Face~\citep{wolf-etal-2020-transformers}, respectively. These two platforms support most of the common NLP models, thus facilitating extension studies of more models in future. We use three common binary text classification datasets --- IMDB movie reviews (IMDB)~\citep{pang2005seeing}, Yelp polarity reviews (YELP)~\citep{zhang2015character}, Quora Question Pair (QQP)~\citep{WinNT} --- as our testbeds. IMDB and YELP datasets present the task of sentiment analysis, where each sentence is labelled as positive or negative sentiment. QQP is a paraphrase detection task, where each pair of sentences is marked as semantically equivalent or not. To control the effect of dataset size and imbalanced classes, all datasets are randomly subsampled to the same size as IMDB (50k) with balanced classes. The training steps for all experiments are the same as well. We implement perturbations $g(\cdot)$ with two self-designed ones and six selected ones from the NL-Augmenter library \citep{dhole2021nl}. For perturbation probabilities, we choose 0.001, 0.005, 0.01, 0.02, 0.05, 0.10, 0.50, 1.00. We run all experiments across three random seeds and report the average results.

\subsection{Perturbation Learnability Analysis}\label{sec:analysis}

Figure~\ref{fig:sensitivity} shows learnability as a function of perturbation probability. Learnability @ $p$ generally increases as we increase the perturbation probability, and when we perturb all the examples (i.e., $p=1.0$), every model can easily identify it well, resulting in the maximum learnability of 1.0. This shows that neural NLP models master these perturbations eventually. At lower perturbation probabilities, some models still learn that perturbation alone predicts the label. In fact, the major difference between different $p-\text{learnability}$ curves is the area of lower perturbation probabilities and this provides motivation for using $\log AUC$ instead of $AUC$ as the summarization of learnability at different $p$ (Section~\ref{sec:defi}).

Table~\ref{tab:benchmark} shows the average learnability over all perturbation probabilities of each model--perturbation pair on IMDB dataset in Figure~\ref{fig:sensitivity}.\footnote{Please refer to Appendix \ref{sec:vis} for benchmark results on YELP (Table \ref{tab:yelp-benchmark}) and QQP (Table \ref{tab:qqp-benchmark}) datasets.} 
% Model-wise comparison\footnote{We note that model-wise comparison is not fair across models with different numbers of parameters. Nevertheless, it is still instructive to compare models at their commonly-used size.} (across different columns in Table~\ref{tab:benchmark}) shows that the non-pretrained model (TextRNN) is generally better at identifying perturbation than pretrained models (BERT, RoBERTa, XLNet). Our results are in line with recent findings that pretraining improves robustness to spurious correlations~\citep{hendrycks2019using,hendrycks-etal-2020-pretrained,tu2020empirical}. We also observe that RoBERTa is less sensitive than BERT, indicating that a larger pretraining corpus improves downstream robustness and confirms that RoBERTa is indeed robustly optimized~\citep{liu2019roberta}. Interestingly, the sensitivity of RoBERTa jumps from 0.0 to 1.0 quickly at a relatively small injection probability, instead of changing gradually as the injection probability increases. This shows that RoBERTa is good at generalizing from a small proportion of data with spurious feature.~\citet{tu2020empirical} also present a similar finding that RoBERTa generalize better from minority patterns in the training set than BERT. We find that learnability complements to existing literature on model interpretations, providing a new perspective and a promising analysis tool.
It reveals the most learnable perturbation for each model. For example, the learnability of  ``visual\_attack\_letters'' and ``leet\_letters'' are very high for all four models, likely due to their strong effects on the tokenization process~\citep{salesky2021robust}.
Perturbations like ``white\_space\_perturbation'' and ``duplicate\_punctuations'' are less learnable for pretrained models, probably because they have weaker 
effects on the subword level tokenization, or they may have encountered similar noise in the pretraining corpora. We observe that ``duplicate\_punctuations'' already exists in the original text of YELP dataset (e.g., ``\textit{The burgers are awesome\textbf{!!}}''), thus violating our assumptions for perturbations in Section \ref{sec:pert}. As a result, the curve for this perturbation substantially deviates from others in Figure~\ref{fig:sensitivity}. We do not count this perturbation on YELP dataset in the following analysis.
% The rank of feature sensitivity differs a lot between models, indicating that a potential solution to a single spurious feature may not work for all models. Priority matters when dealing with spurious correlations.
The perturbation learnability experiments provide a clean setup for NLP practitioners to analyze the effect of textual perturbations on models.

\begin{table}\centering
\begin{tabular}{lccc}\toprule
$\rho$ &IMDB &YELP &QQP \\\cmidrule{1-4}
\makecell[l]{Avg. learnability \\vs. robustness} &-0.643* &-0.821* &-0.695* \\\cmidrule{1-4}
\makecell[l]{Avg. learnability \\vs. post aug $\Delta$} &0.756* &0.846* &0.750* \\
\bottomrule
\end{tabular}
\caption{Correlations of average learnability vs. robustness vs. post data augmentation $\Delta$. $\rho$ is Spearman correlation. $^{*}$ indicates high significance (p-value $<$ 0.001). }\label{tab:corr}
\end{table}

\subsection{Empirical Findings}\label{sec:finding}
We observe a negative correlation between learnability (Equation~\ref{eq:learnability}) and robustness (Equation~\ref{eq:robustness}) across all three datasets in Table~\ref{tab:benchmark}, validating Hypothesis~\ref{hyp:robustness}. Table~\ref{tab:benchmark} also quantifies the trend that data augmentation with a perturbation the model is \textit{less} robust to has \textit{more} improvement on robustness (Hypothesis~\ref{hyp:data-aug}). We plot the correlations on IMDB dataset in Figure~\ref{fig:imdb-robustness} and~\ref{fig:imdb-data-aug}.\footnote{For visualizations of correlations on the other two datasets, please refer to Figure \ref{fig:yelp-corr} for YELP and Figure \ref{fig:qqp-corr} for QQP in Appendix \ref{sec:vis}.} Both the correlations between 1) learnability vs. robustness and 2) learnability vs. improvement by data augmentation are strong (Spearman $\abs{\rho}>0.6$) and highly significant (p-value $<$ 0.001), which firmly supports our hypotheses. Our findings provide insight about when the model is less robust and when data augmentation works better for improving robustness.

Figure~\ref{fig:imdb-robustness-data-aug-sensitivity} shows that the more learnable a perturbation is for a model, the greater the likelihood that its robustness can be improved through data augmentation along this perturbation. We argue that this is not simply because there is more room for improvement by data augmentation. From a causal perspective, learnability acts as a common cause (confounder) for both robustness and improvement by data augmentation. This indicates a potential limitation of using data augmentation for improving robustness to perturbations~\citep{jha2020does}:  
data augmentation is \textit{only} more effective at improving robustness against perturbations more learnable for a model.
% for unlearnable perturbations, data augmentation may be of little help 
% Approaches that go beyond simple data augmentation are required to combat such perturbations. 

\section{Discussion}\label{sec:discussion}

\paragraph{Potential Impacts.}
Our findings seem intuitive but are non-trivial. The NLP models were not trained on perturbed examples when measuring robustness, but still they display a strong correlation with perturbation learnability. Understanding these findings are important for a more principled evaluation of and control over NLP models~\citep{lovering2020predicting}. Specifically, the learnability metric complements to the evaluation of newly designed perturbations by revealing model weaknesses in a clean setup. Reducing perturbation learnability is promising for improving robustness of models. Contrastive learning~\citep{gao2021simcse,yan2021consert} that pulls the representations of the original and perturbed text together, makes it difficult for the model to identify the perturbation (reducing learnability) and thus may help improve robustness. 
Perturbation can also be viewed as injecting \textit{spurious} feature into the examples, so the learnability metric also helps to interpret robustness to spurious correlation~\citep{sagawa2020investigation}. 
Moreover, learnability may facilitate the development of model architectures with explicit inductive biases~\citep{warstadt2020can,lovering2020predicting} to avoid sensitivity to noisy perturbations. Grounding the learnability within the causality framework inspires future researchers to incorporate the causal perspective into model design~\cite{zhang2020causal}, and make the model robust to different types of perturbations.

\paragraph{Limitations.}
In this work, we focus on the robust accuracy (Section~\ref{sec:defi}), which is accuracy on the perturbed test set. We do not assume that the test accuracy of the original test set, a.k.a in-distribution accuracy, is invariant  invariant against training with augmentation or not. It would be interesting to investigate the trade-off between robust accuracy and in-distribution accuracy in the future.
We also note that this work has not established that the relationship between learnability and robustness is \textit{causal}. This could be explored with other approaches in causal inference for deconfounding besides simulation on randomized control trial, such as working with real data but stratifying it~\citep{frangakis2002principal}, to bring the learnability experiment closer to more naturalistic settings. Although we restrict to balanced, binary classification for simplicity in this pilot study, our framework can also be extended to imbalanced, multi-class classification. 
We are aware that computing average learnability is  expensive for large models and datasets, which is further discussed in Section \ref{sec:ethics}. We provide a greener solution in Appendix \ref{sec:learnability-p}.
We could further verify our assumptions for perturbations with a user study~\citep{moradi2021evaluating} which investigates how understandable the perturbed texts are to humans. 

\section{Related Work}\label{sec:related}

\paragraph{Robustness of NLP Models to Perturbations.}
The performance of NLP models can decrease when encountering noisy data in the real world. Recent works~\cite{prabhakaran2019perturbation,ribeiro2020beyond,niu2020evaluating,moradi2021evaluating} present comprehensive evaluations of the robustness of NLP models to different types of perturbations, including typos, changed entities, negation, etc. Their results reveal the phenomenon that NLP models can handle some specific types of perturbation more effectively than others. However, they do not go into a deeper analysis of the reason behind the difference of robustness between models and perturbations.

% \paragraph{Training with Random Labels.}~\citet{pondenkandath2018leveraging,maennel2020neural,zhang2021understanding} trained deep neural networks (DNNs) on Computer Vision (CV) datasets with entirely random labels to study  memorization, generalization, pretraining, and alignment. Though we similarly use random label assignment (Section~\ref{sec:random}) , our work is different from previous work in that 1) our insights behind randomization  originate from the concept of Randomized Controlled Trial (RCT) in causality; 2) we instead use randomization to study learnability of textual perturbations in NLP; 3) our labels are not purely random: they are correlated with the existence of perturbations.

\paragraph{Interpretation of Data Augmentation.}
Although data augmentation has been widely used in CV~\citep{sato2015apac,devries2017improved,dwibedi2017cut} and NLP~\citep{wang-yang-2015-thats,kobayashi2018contextual,wei2019eda},  the underlying mechanism of its effectiveness remains under-researched.  Recent studies aim to quantify intuitions of how data augmentation improves model generalization.~\citet{gontijo2020tradeoffs} introduce affinity and diversity, and find a correlation between the two metrics and augmentation performance in image classification.  In NLP,~\citet{kashefi2020quantifying} propose a KL-divergence--based metric to predict augmentation performance. Our proposed learnability metric implies when data augmentation works better and thus acts as a complement to this line of research.

\section{Conclusion}\label{sec:conclusion}
This work targets at an open question in NLP: why models are less robust to some textual perturbations than others? We find that learnability, which causally quantifies how well a model learns to identify a perturbation, is predictive of the model robustness to the perturbation.  
In future work, we will investigate whether these findings can generalize to other domains, including computer vision.
% , and look at their implications for robust training and causality-aware modelling methods.

\section{Ethics Statement}\label{sec:ethics}

Computing average learnability requires training a model for multiple times at different perturbation probabilities, which can be computationally intensive if the sizes of the datasets and models are large. This can be a non-trivial problem for NLP practitioners with limited computational resources. We hope that our benchmark results of typical perturbations for NLP models work as a reference for potential users. Collaboratively sharing the results of such metrics on popular models and perturbations in public fora can also help reduce duplicate investigation and coordinate efforts across teams.

To alleviate the computational efficiency issue of average learnability estimation, using learnability at selected perturbation probabilities may help at the cost of reduced precision (Appendix~\ref{sec:learnability-p}). We are not alone in facing this issue: two similar metrics for interpreting model inductive bias, \textit{extractability} and \textit{s-only error}~\citep{lovering2020predicting} also require training the model repeatedly over the whole dataset. Therefore, finding an efficient proxy for average learnability is promising for more practical use of learnability in model interpretation.

\section*{Acknowledgements}
This research is supported by the National Research Foundation, Singapore under its International Research Centres in Singapore Funding Initiative. Any opinions, findings and conclusions or recommendations expressed in this material are those of the author(s) and do not reflect the views of National Research Foundation, Singapore. We acknowledge the support of NVIDIA Corporation for their donation of the GeForce RTX 3090 GPU that facilitated this research.

\bibliography{anthology,custom}
\bibliographystyle{acl_natbib}

\clearpage
\appendix
\section{Algorithm for Perturbation Learnability Estimation}\label{sec:algo}

\begin{algorithm}[H]
\caption{Learnability Estimation}\label{alg:algorithm}
% \begin{multicols}{2}
\textbf{Input}: training set $D_{train}=\{(x_{1}, l_{1}), ...,  (x_{n}, l_{n})\}$, test set $D_{test}=\{(x_{n+1}, l_{n+1}), ...,  (x_{n+m}, l_{n+m})\}$, $D=D_{train} \cup D_{test}$, model $f : (x;\theta) \mapsto \{0,1\}$, perturbation $g : (x;\beta) \rightarrow x^{*} $, perturbation probability $p$\\
\textbf{Output}: $\text{learnability}(f,g,p,D)$
\begin{algorithmic}[1] 
\STATE // \ding{172} assigning random labels
\STATE Initialize an empty dataset  $D^{\prime}$
\FOR{$i$ in $\{1,2,...,n+m\}$}  
\STATE $l_{i}^{\prime} \leftarrow randint[0,1]$
\STATE $D^{\prime} \leftarrow D^{\prime} \cup \{(x_{i}, l_{i}^{\prime})\}$ 
\ENDFOR
\STATE // \ding{173} perturbing  with  probabilities
\STATE Initialize an empty dataset $D^{\prime*}$
\FOR{$i$ in $\{1,2,...,n+m\}$}
\STATE $z \leftarrow rand(0,1)$

\STATE $x_{i}^{*} \leftarrow x_{i}$
\IF{$l_{i}^{\prime}=1 \wedge z<p$}
\STATE $x_{i}^{*} \leftarrow g(x_{i})$

\ENDIF
\STATE $D^{\prime *} \leftarrow D^{\prime *} \cup \{(x_{i}^{*}, l_{i}^{\prime})\}$ 
\ENDFOR
\STATE // \ding{174} estimating model performance
\STATE $D_{train}^{\prime},D_{test}^{\prime} \leftarrow D^{\prime}[1:n], D^{\prime}[n+1:n+m]$
\STATE $D_{train}^{\prime *},D_{test}^{\prime *} \leftarrow D^{\prime *}[1:n],D^{\prime *}[n+1:n+m]$
\STATE fit the model $f(\cdot)$ on $D_{train}^{\prime *}$
\STATE $\mathcal{A}(f,g,p,D_{test}^{\prime *}) \leftarrow f(\cdot)$ accuracy on $D_{test}^{\prime *}$
\STATE $\mathcal{A}(f,g,p,D_{test}^{\prime}) \leftarrow f(\cdot)$ accuracy on $D_{test}^{\prime}$
\STATE \textbf{return} $\mathcal{A}(f,g,p,D_{test}^{\prime *}) - \mathcal{A}(f,g,p,D_{test}^{\prime}) $
\end{algorithmic}
% \end{multicols}
\end{algorithm}

\section{Background on Causal Inference}\label{sec:background}

The aim of causal inference is to investigate how a treatment $T$ affects the outcome $Y$. Confounder $X$ refers to a variable that influences both treatment $T$ and outcome $Y$.  For example,  sleeping with shoes on ($T$) is strongly associated with waking up with a headache ($Y$), but they both have a common cause: drinking the night before ($X$)~\citep{neal2020introduction}. In our work, we aim to study how a perturbation (treatment) affects the model's prediction (outcome). However, the latent features and other noise usually act as confounders.

Causality offers solutions for two questions: 1) how to eliminate the spurious association and isolate the treatment's causal effect; and 2) how varying $T$ affects $Y$, given both variables are causally-related~\citep{liu2021everything}. We leverage both of these properties in our proposed method.  Let us now introduce Randomized Controlled Trial and Average Treatment Effect as key concepts in answering the above two questions, respectively.

\paragraph{Randomized Controlled Trial (RCT).}
 In an RCT, each participant is randomly assigned to either the treatment group or the non-treatment group. In this way, the only difference between the two groups is the treatment they receive. Randomized experiments ideally guarantee that there is no confounding factor, and thus any observed association is actually causal. We operationalize RCT as a perturbation classification task in Section~\ref{sec:random}. 

\paragraph{Average Treatment Effect (ATE).}
In Section~\ref{sec:identify}, we apply ATE~\citep{holland1986statistics} as a measure
of learnability. ATE is based on Individual Treatment Effect (ITE, Equation~\ref{eq:ITE}), which is the difference of the outcome with and without treatment.
\begin{equation}
ITE_{i}=Y_{i}(1)-Y_{i}(0). \label{eq:ITE}
\end{equation}
Here, $Y_{i}(1)$ is the outcome $Y$ of individual $i$ that receives treatment ($T=1$), while $Y_{i}(0)$ is the opposite. In the above example, waking up with a headache ($Y=1$) with shoes on ($T=1$) means  $Y_{i}(1)=1$.

We calculate the Average Treatment Effect (ATE) by taking an average over ITEs:
\begin{equation}
    ATE=\mathbb{E}[Y(1)]-\mathbb{E}[Y(0)]. \label{eq:ATE}
\end{equation}
ATE quantifies how the outcome $Y$ is expected to change if we modify the treatment $T$ from 0 to 1. We provide specific definitions of ITE and ATE in Section~\ref{sec:identify}.

\section{Alternate Definition of Perturbation Learnability}\label{sec:alter}

In Section~\ref{sec:identify}, we propose an accuracy-based identification of ATE. Now we discuss another probability-based identification and compare between them. We can also define the outcome $Y$ of a test example $x_{i}$ as the predicted probability of (pseudo) true label given by the trained model $f(\cdot)$:
\begin{equation}
    Y_{i}(0)\coloneqq P_{f}(L^{\prime}=l_{i}^{\prime}\mid X=x_{i}) \in (0,1).
\label{eq:Y0-prob}
\end{equation}
Similarly, the performance outcome $Y$ of a perturbed test data point $x_{i}^{*}$ is:
\begin{equation}
    Y_{i}(1)\coloneqq P_{f}(L^{\prime}=l_{i}^{\prime}\mid X=x_{i}^{*}) \in (0,1).
\label{eq:Y1-prob}
\end{equation}
For example, for a test example $(x_{i},l_{i}^{\prime})$ which receives treatment ($l_{i}^{\prime}=1$), the trained model $f(\cdot)$ predicts its label as 1 with only a small probability 0.1 before treatment (it has not been perturbed yet), and 0.9 after treatment. So the Individual Treatment Effect (ITE, see Equation~\ref{eq:ITE}) of this example is calculated as $ITE_{i}=Y_{i}(1)-Y_{i}(0)=0.9-0.1=0.8$. We then take an average over all the perturbed test examples (half of the test set)
% \footnote{The other half of the test set ($l^{\prime}=0$) is left unperturbed, following the same procedure in Section~\ref{sec:defi}. Therefore, we do not take them into account for ATE calculation.}
as Average Treatment Effect (ATE, see Equation~\ref{eq:ATE}), which is exactly the learnability of a perturbation for a model. To clarify, the two operands in Equation~\ref{eq:ATE} are defined as follows:
\begin{equation}
    \mathbb{E}[Y(1)] \coloneqq \mathcal{P}(f,g,p,D_{test}^{\prime *}).
    \label{eq:EY1}
\end{equation}
It means the average predicted probability of (pseudo) true label given by the trained model $f(\cdot)$ on the perturbed test set $D_{test}^{\prime *}$.
\begin{equation}
    \mathbb{E}[Y(0)] \coloneqq \mathcal{P}(f,g,p,D_{test}^{\prime}).  
\label{eq:EY0}
\end{equation}
Similarly, this is the average predicted probability on the randomly labeled test set $D_{test}^{\prime}$.

Notice that the accuracy-based definition of outcome $Y$ (Equation~\ref{eq:Y0-accu}) can also be written in a similar form to the probability-based one (Equation~\ref{eq:Y0-prob}):
\begin{equation}
    Y_{i}(0)\coloneqq \mathbf{1}_{\{f(x_{i})=l_{i}^{\prime}\}} = \mathbf{1}_{\{P_{f}(L^{\prime}=l_{i}^{\prime}\mid X=x_{i})>0.5\}} \in \{0,1\}.
\end{equation}
because the correctness of the prediction is equal to whether the predicted probability of true (pseudo) label exceeds a certain threshold (i.e., 0.5). 

The major difference is that, accuracy-based $ITE$  is a discrete variable falling in $\{-1,0,1\}$, while probability-based $ITE$ is a continuous one ranging from -1 to 1. For example, if a model learns to identify a perturbation and thus changes its prediction from wrong (before perturbation) to correct (after perturbation), accuracy-based $ITE$ will be $1-0=1$ while probability-based $ITE$ will be less than 1. That is to say, accuracy-based $ATE$ tends to vary more drastically than probability-based if inconsistent predictions occur more often, and thus can better capture the nuance of perturbation learnability. 
Empirically, we find that accuracy-based average learnability varies greatly ($\sigma=0.375$, Table~\ref{tab:sensitivity-p}) and thus can better distinguish between different model-perturbation pairs than probability-based one ($\sigma=0.288$, Table~\ref{tab:sensitivity-p}). As a result, we choose accuracy-based ATE as the primary measurement of learnability in this paper.

\begin{table*}
\centering
\begin{tabular}{l|cccc|cccc}\toprule
\multirow{2}{*}{$p$} &\multicolumn{4}{c|}{Accuracy-based Learnability @ $p$} &\multicolumn{4}{c}{Probability-based Learnability @ $p$} \\\cmidrule{2-9}
&$\sigma$ &Avg Learn. &Robu. &Post Aug $\Delta$ &$\sigma$ &Avg Learn. &Robu. &Post Aug $\Delta$ \\\cmidrule{1-9}
Avg. &0.375 &1.000* &-0.643* &0.756* &0.288 &1.000* &-0.652* &0.727* \\\cmidrule{1-9}
0.001 &0.182 &0.426* &-0.265 &0.259 &0.114 &0.367* &-0.279 &0.288 \\
0.005 &0.235 &0.637* &-0.383* &0.522* &0.192 &0.925* &-0.620* &0.702* \\
0.01 &0.263 &0.741* &-0.530* &0.635* &0.192 &0.893* &-0.567* &0.586* \\
0.02 &0.257 &0.816* &-0.636* &0.743* &0.192 &0.886* &-0.686* &0.690* \\
0.05 &0.236 &0.279 &-0.158 &0.136 &0.121 &0.576* &-0.371* &0.350* \\
0.1 &0.241 &0.354* &-0.162 &0.192 &0.115 &0.543* &-0.288 &0.258 \\
0.5 &0.094 &0.024 &0.155 &-0.179 &0.037 &-0.080 &0.114 &-0.258 \\
1.0 &0.011 &-0.199 &0.252 &-0.332 &0.019 &-0.220 &0.294 &-0.402* \\
\bottomrule
\end{tabular}
\caption{Standard deviations ($\sigma$) of Learnability @ $p$ and  Spearman correlations between accuracy-based/probability-based learnability @ $p$ vs. average learnability/robustness/post data augmentation $\Delta$ over all model-perturbation pairs on IMDB dataset. $^{*}$ indicates significance (p-value $<$ 0.05).}\label{tab:sensitivity-p}
\end{table*}

\section{Investigating Learnability at a Specific Perturbation Probability}\label{sec:learnability-p}

Inspired by Precision @ K in Information Retrieval (IR), we propose a similar metric dubbed Learnability @ $p$, which is the learnability of a perturbation for a model  at a specific perturbation probability $p$. We are primarily interested in whether a selected $p$ can represent the learnability over different perturbation probabilities and correlates well with robustness and post data augmentation $\Delta$. 

We calculate the standard deviation ($\sigma$) of Learnability @ $p$ and average learnability ($\log AUC$) over all model-perturbation pairs to measure how well it can distinguish between different models and perturbations. Table~\ref{tab:sensitivity-p} shows that average learnability is more diversified than all Learnability @ $p$ and diversity ($\sigma$) peaks at $p=0.01$ for accuracy-based/probability-based measurement. Accuracy-based Learnability @ $p$ is generally more diversified across models and perturbations than its counterpart. 
To investigate the strength of the correlations, we also calculate Spearman $\rho$ between accuracy-based/probability-based learnability @ $p$ vs. average learnability/robustness/post data augmentation $\Delta$ over all model-perturbation pairs. Table~\ref{tab:sensitivity-p} shows that generally average learnability has stronger correlation than Learnability @ $p$. Correlations with both robustness and post data augmentation $\Delta$ peak at $p=0.02$ for accuracy-based/probability-based measurements, and the correlations with average learnability (0.816*/0.886*) are also strong at these perturbation probabilities. 

Overall, Learnability @ $p$ with higher standard deviation correlates better with average learnability, robustness and post data augmentation $\Delta$. Our analysis shows that if $p$ is carefully selected by $\sigma$, Learnability @ $p$ is also a promising metric, though not as accurate as average learnability. One advantage of Learnability @ $p$ over average learnability is that it costs less time to obtain learnability at a single perturbation probability. 
% We plan to explore other efficient proxies of average learnability in future. 

\section{Additional Experiment Results}\label{sec:vis}

\begin{figure*}
     \centering
     \begin{subfigure}[b]{0.27\textwidth}
         \centering
         \includegraphics[width=\textwidth]{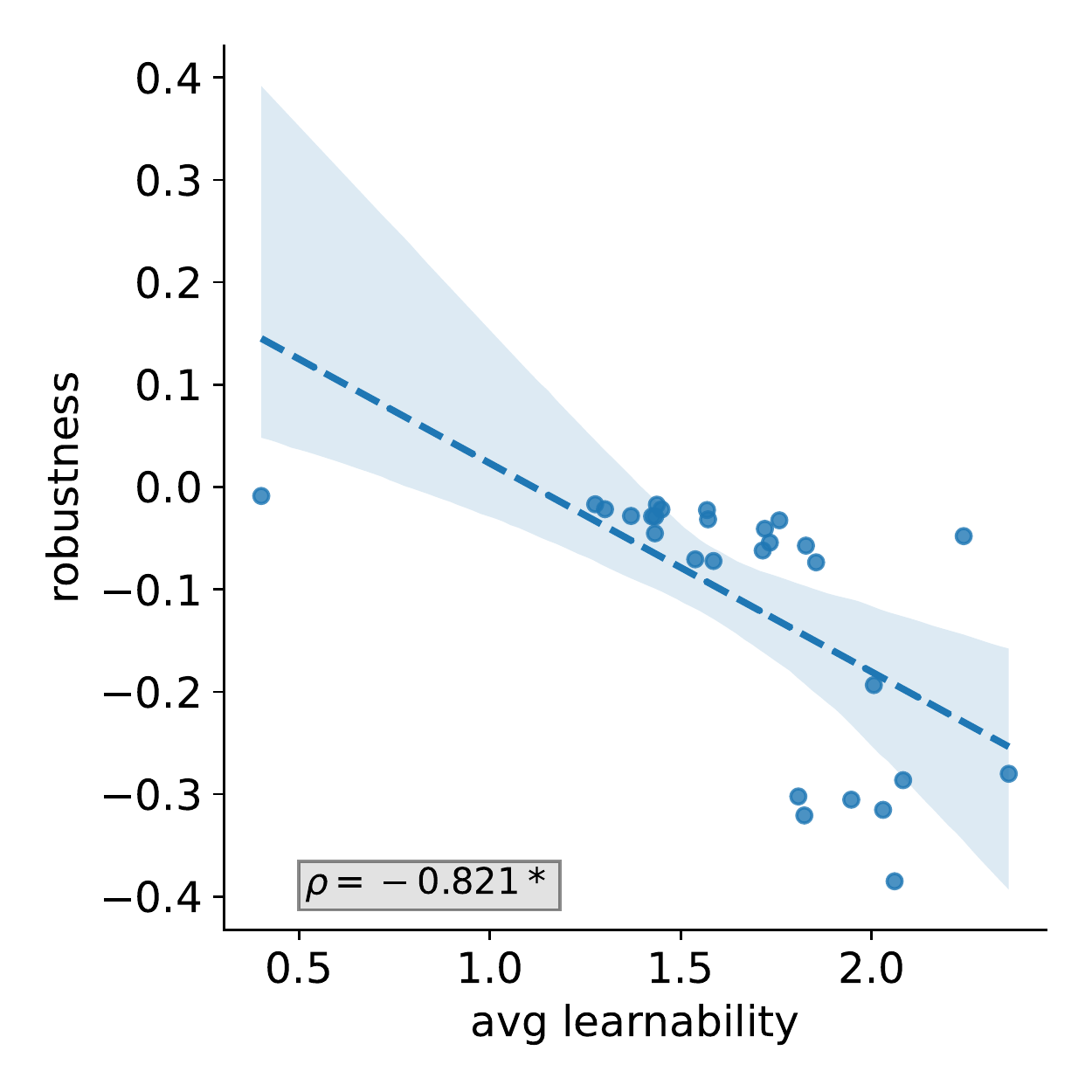}
         \caption{Learnability vs. Robustness}
         \label{fig:yelp-robustness}
     \end{subfigure}
     \hfill
     \begin{subfigure}[b]{0.27\textwidth}
         \centering
         \includegraphics[width=\textwidth]{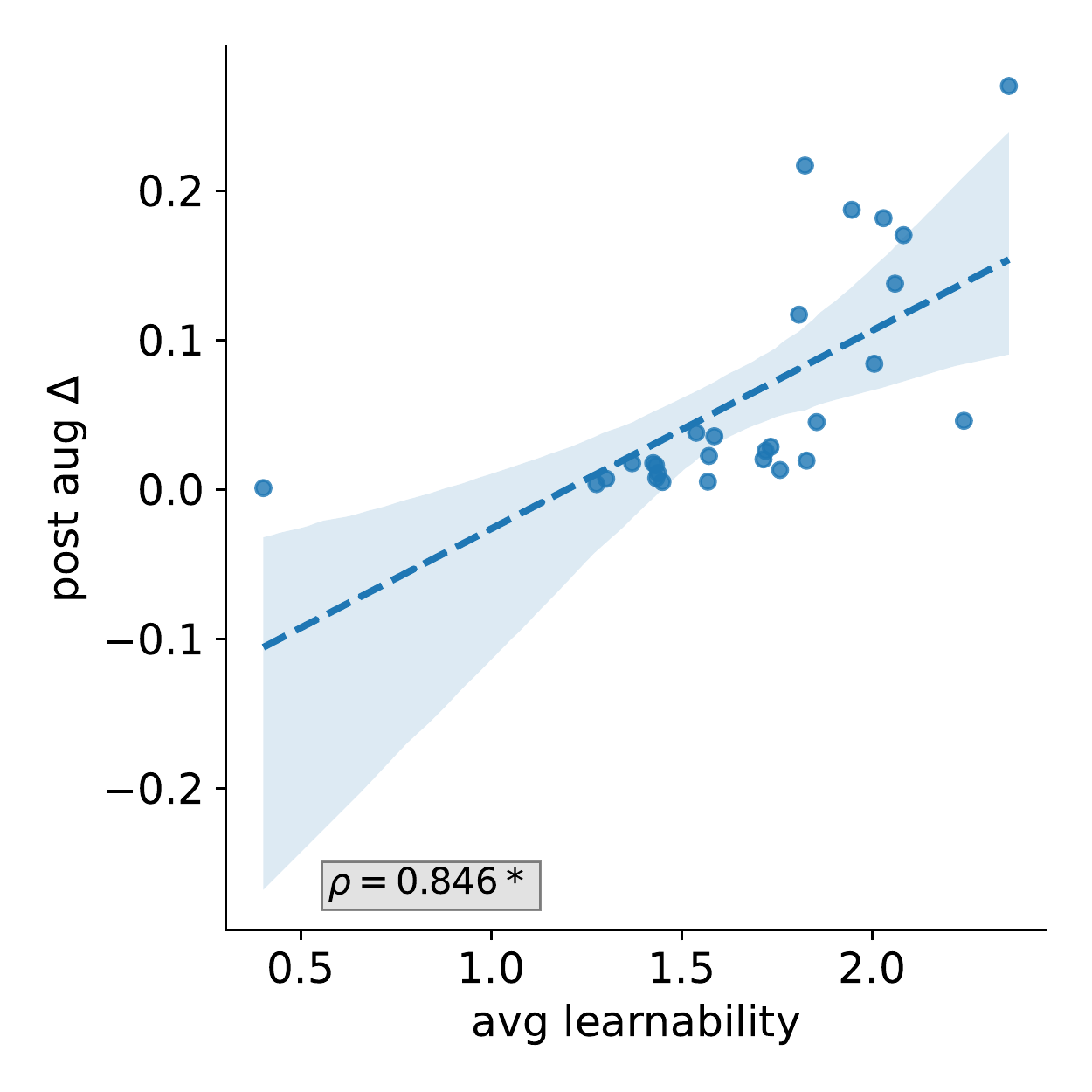}
         \caption{Learnability vs. Post Aug $\Delta$}
         \label{fig:yelp-data-aug}
     \end{subfigure}
     \hfill
     \begin{subfigure}[b]{0.38\textwidth}
         \centering
         \includegraphics[width=\textwidth]{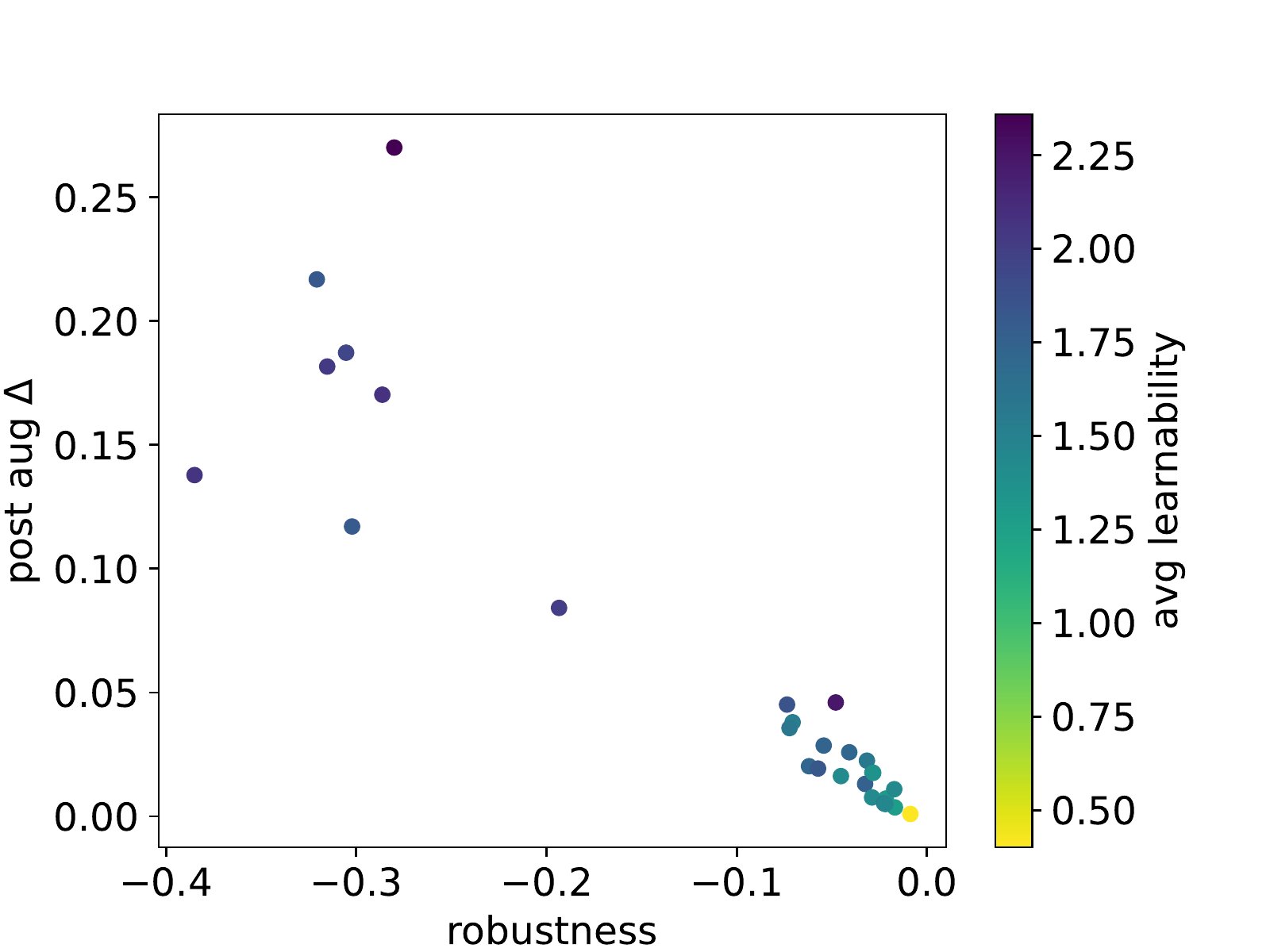}
         \caption{Learn. vs. Robu. vs. Post Aug $\Delta$}
         \label{fig:yelp-robustness-data-aug-sensitivity}
     \end{subfigure}
        \caption{Linear regression plots of learnability vs. robustness vs. post data augmentation $\Delta$  on \textbf{YELP} dataset. Each point in the plots represents a model-perturbation pair. $\rho$ is Spearman correlation. $^{*}$ indicates high significance (p-value $<$ 0.001).   }
        \label{fig:yelp-corr}
\end{figure*}

\begin{figure*}
     \centering
     \begin{subfigure}[b]{0.27\textwidth}
         \centering
         \includegraphics[width=\textwidth]{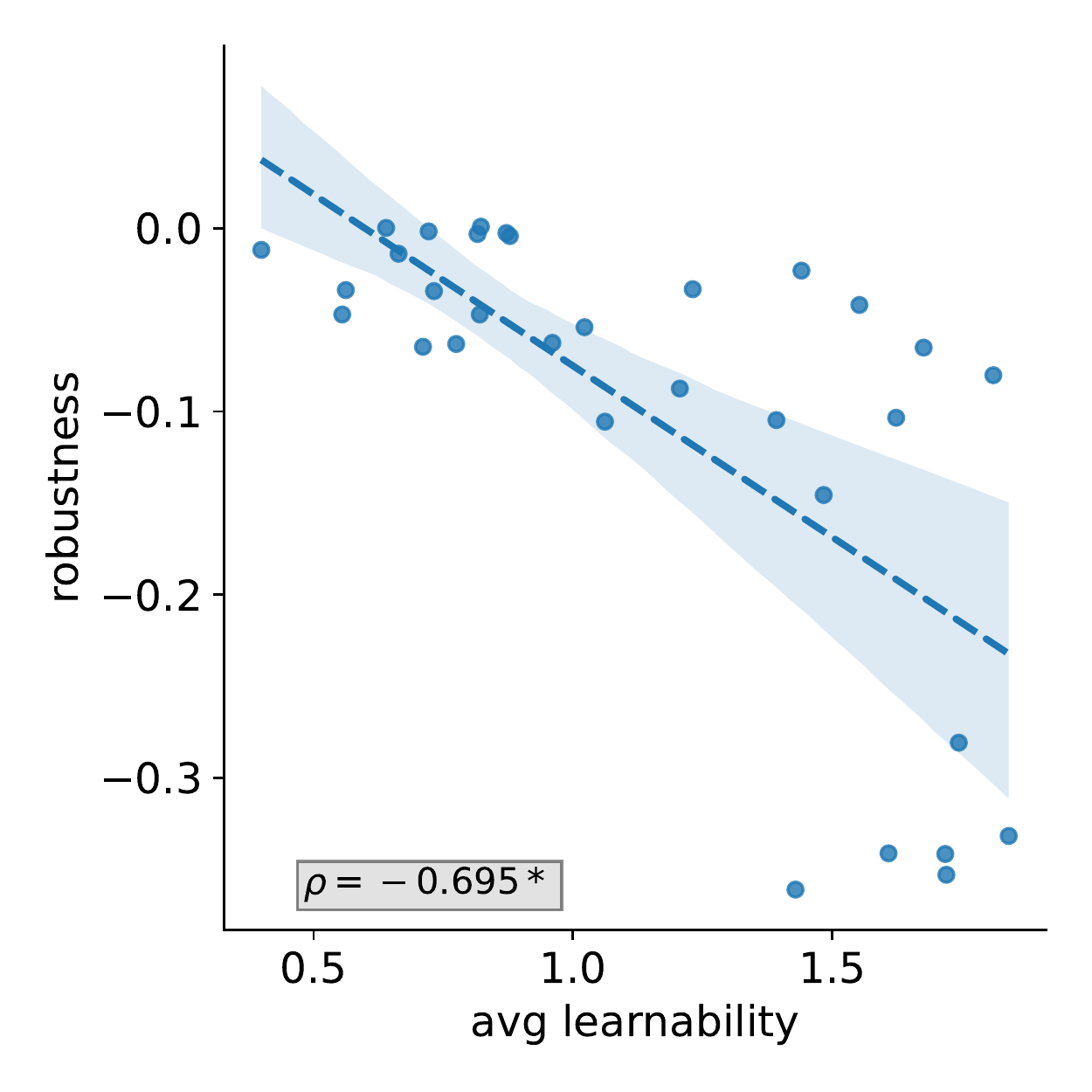}
         \caption{Learnability vs. Robustness}
         \label{fig:qqp-robustness}
     \end{subfigure}
     \hfill
     \begin{subfigure}[b]{0.27\textwidth}
         \centering
         \includegraphics[width=\textwidth]{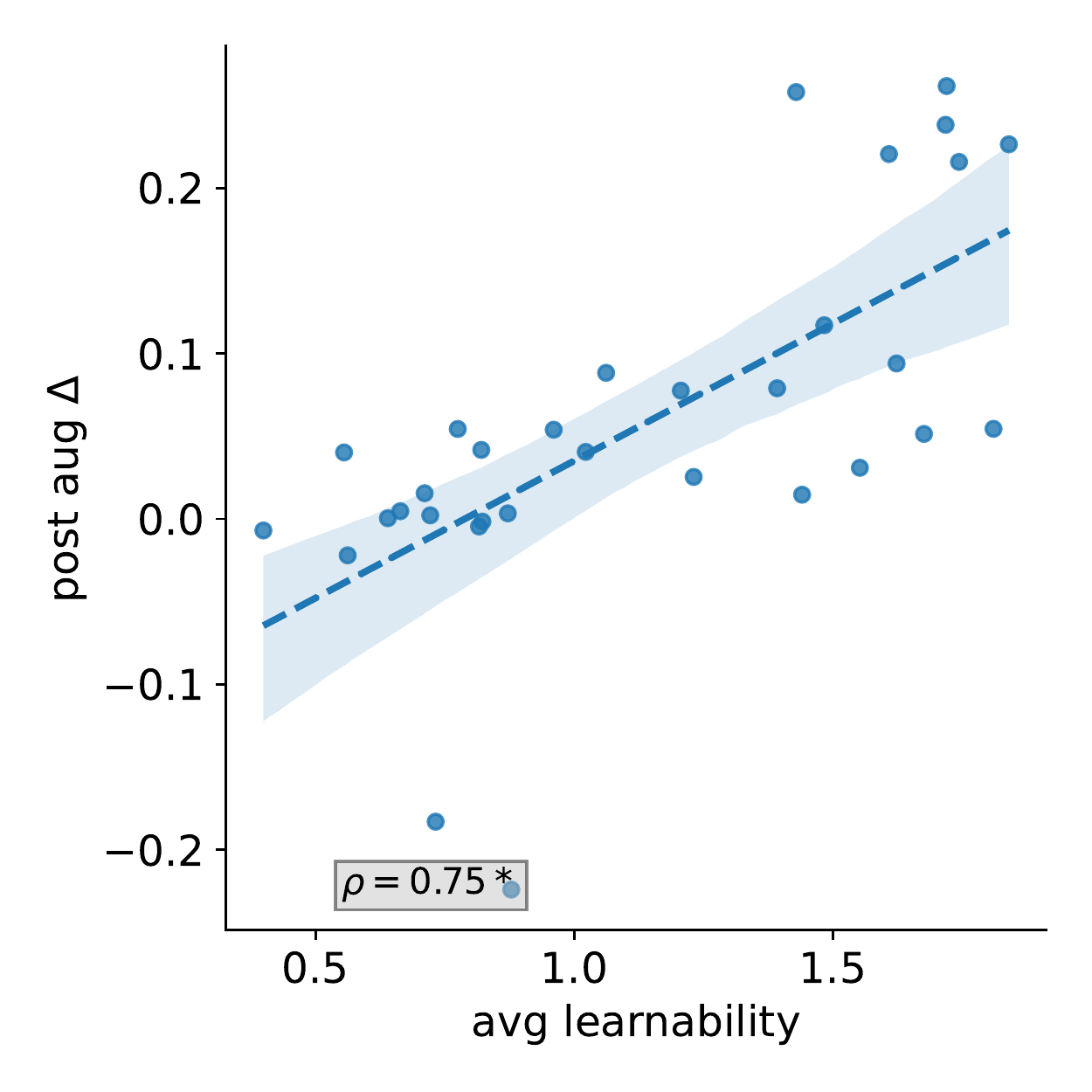}
         \caption{Learnability vs. Post Aug $\Delta$}
         \label{fig:qqp-data-aug}
     \end{subfigure}
     \hfill
     \begin{subfigure}[b]{0.38\textwidth}
         \centering
         \includegraphics[width=\textwidth]{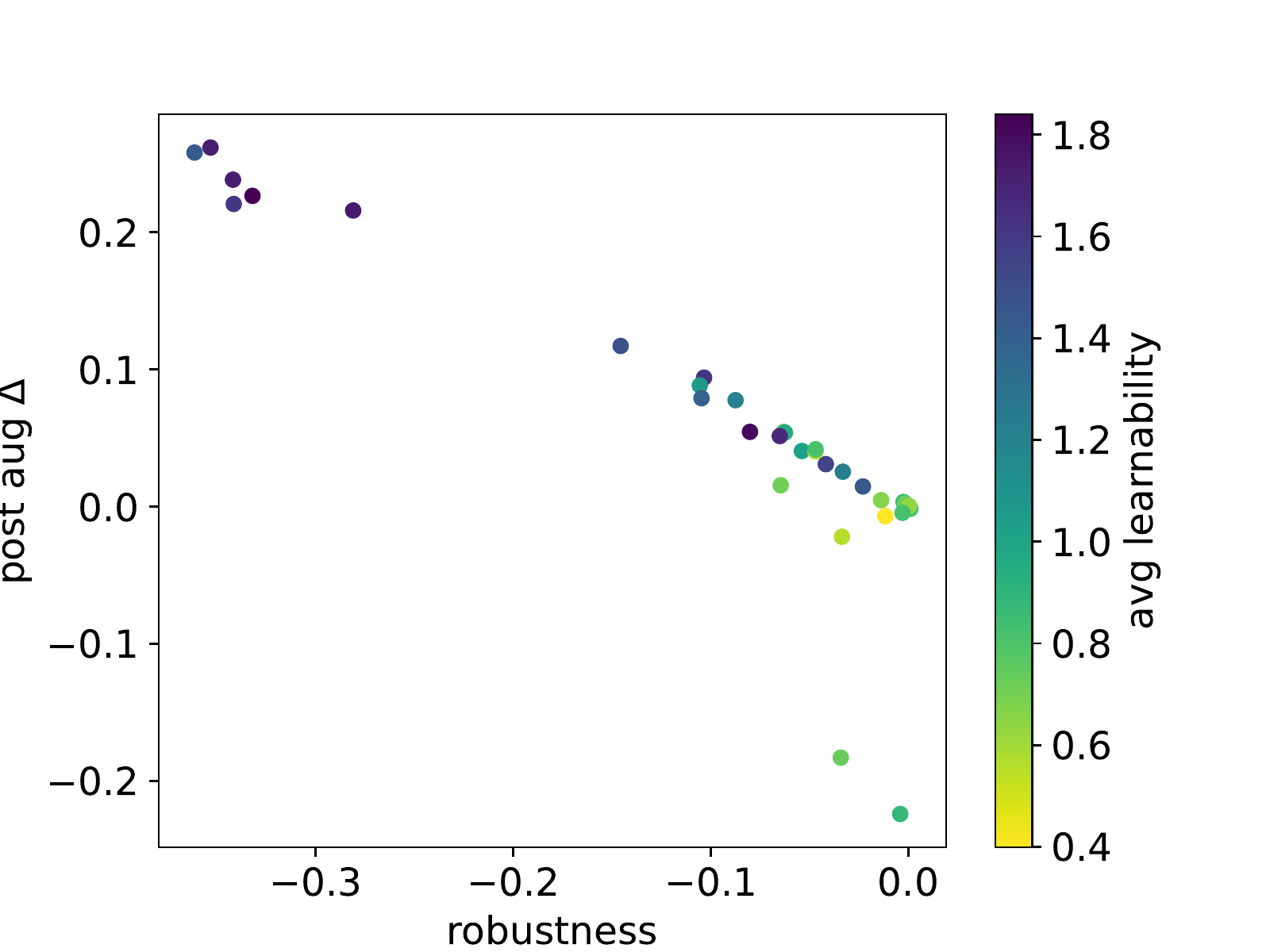}
         \caption{Learn. vs. Robu. vs. Post Aug $\Delta$}
         \label{fig:qqp-robustness-data-aug-sensitivity}
     \end{subfigure}
        \caption{Linear regression plots of learnability vs. robustness vs. post data augmentation $\Delta$  on \textbf{QQP} dataset. Each point in the plots represents a model-perturbation pair. $\rho$ is Spearman correlation. $^{*}$ indicates high significance (p-value $<$ 0.001).   }
        \label{fig:qqp-corr}
\end{figure*}

\begin{table*}
\centering
% Table generated by Excel2LaTeX from sheet 'benchmark_yelp_small'
\begin{tabular}{l|cccc|c}
\toprule
Perturbation & RoBERTa & XLNet & TextRNN & BERT  & \makecell[c]{Average\\over models} \\
\midrule
shuffle\_word & 1.538 & 1.586 & 0.401 & \textbf{1.854} & 1.345 \\
butter\_fingers\_perturbation & 1.301 & 1.433 & 1.425 & 1.758 & 1.479 \\
whitespace\_perturbation & 1.276 & 1.449 & 1.720  & 1.569 & 1.504 \\
insert\_abbreviation & 1.437 & 1.370  & \underline{2.241} & 1.572 & 1.655 \\
random\_upper\_transformation & 1.432 & 1.828 & 1.733 & 1.715 & 1.677 \\
visual\_attack\_letters & \underline{2.060}  & \textbf{2.006} & 2.030  & 1.808 & \underline{1.976} \\
leet\_letters & \textbf{2.083} &	\underline{1.947} &	\textbf{2.359} &	\underline{1.824} &	\textbf{2.053} \\
\bottomrule
\end{tabular}%
   \caption{Average learnability ($\log AUC$ of corresponding curve in Figure~\ref{fig:sensitivity}) of each model--perturbation pair on YELP dataset. Rows are sorted by average values over all models. The perturbation for which a model is most learnable is highlighted in \textbf{bold} while the following one is \underline{underlined}.} \label{tab:yelp-benchmark}
\end{table*}

\begin{table*}
\centering
% Table generated by Excel2LaTeX from sheet 'benchmark_qqp_1'
\begin{tabular}{l|cccc|c}
\toprule
Perturbation & RoBERTa & TextRNN & XLNet & BERT  & \makecell[c]{Average\\over models} \\
\midrule
whitespace\_perturbation & 0.732 & 0.399 & 0.562 & 0.711 & 0.601 \\
duplicate\_punctuations & 0.722 & 0.823 & 0.640  & 0.872 & 0.764 \\
butter\_fingers\_perturbation & 0.555 & 0.878 & 0.775 & 1.022 & 0.808 \\
insert\_abbreviation & 0.820  & 1.440  & 0.960  & 1.206 & 1.107 \\
random\_upper\_transformation & 1.062 & 0.664 & 1.392 & 1.483 & 1.150 \\
shuffle\_word & 1.231 & 0.816 & 1.552 & \underline{1.623} & 1.306 \\
visual\_attack\_letters & \underline{1.429} & \textbf{1.810}  & \underline{1.744} & 1.608 & \underline{1.648} \\
leet\_letters & \textbf{1.720}  & \underline{1.676} & \textbf{1.840}  & \textbf{1.718} & \textbf{1.738} \\
\bottomrule
\end{tabular}%
   \caption{Average learnability ($\log AUC$ of corresponding curve in Figure~\ref{fig:sensitivity}) of each model--perturbation pair on QQP dataset. Rows are sorted by average values over all models. The perturbation for which a model is most learnable is highlighted in \textbf{bold} while the following one is \underline{underlined}.} \label{tab:qqp-benchmark}
\end{table*}

\end{document}